\documentclass[10pt,twocolumn,letterpaper]{article}

\usepackage{cvpr}
\usepackage{times}
\usepackage{epsfig}
\usepackage{graphicx}
\usepackage{amsmath}
\usepackage{amssymb}

\usepackage{verbatim}
\usepackage{multirow}
\usepackage{booktabs}
\usepackage{array}

\cvprfinalcopy 

\def\cvprPaperID{1709} 

\ifcvprfinal\fi
\begin{document}

\title{Active Convolution: Learning the Shape of Convolution for Image Classification}

\author{Yunho Jeon\\
EE, KAIST\\
{\tt\small jyh2986@kaist.ac.kr}
\and
Junmo Kim\\
EE, KAIST\\
{\tt\small junmo.kim@kaist.ac.kr}
}

\maketitle

\begin{abstract}
In recent years, deep learning has achieved great success in many computer vision applications. Convolutional neural networks (CNNs) have lately emerged as a major approach to image classification. Most research on CNNs thus far has focused on developing architectures such as the Inception and residual networks. The convolution layer is the core of the CNN, but few studies have addressed the convolution unit itself. In this paper, we introduce a convolution unit called the active convolution unit (ACU). A new convolution has no fixed shape, because of which we can define any form of convolution. Its shape can be learned through backpropagation during training. Our proposed unit has a few advantages. First, the ACU is a generalization of convolution; it can define not only all conventional convolutions, but also convolutions with fractional pixel coordinates. We can freely change the shape of the convolution, which provides greater freedom to form CNN structures. Second, the shape of the convolution is learned while training and there is no need to tune it by hand. Third, the ACU can learn better than a conventional unit, where we obtained the improvement simply by changing the conventional convolution to an ACU. We tested our proposed method on plain and residual networks, and the results showed significant improvement using our method on various datasets and architectures in comparison with the baseline.
\end{abstract}
{\renewcommand{\arraystretch}{1.2}%

\section{Introduction}
Following the success of deep learning in the ImageNet Large Scale Visual Recognition Challenge (ILSVRC)~\cite{ILSVRC15}, the best performance in classification competitions has almost invariably been achieved on convolutional neural network (CNN) architectures. AlexNet~\cite{Krizhevsky2012} is composed of three types of receptive field convolutions ($3 \times 3$, $5 \times 5$, $11 \times 11$). VGG~\cite{Simonyan2015} is based on the idea that a stack of two convolutional layers with a receptive field $3 \times 3$ is more effective than a $5 \times 5$ convolution. GoogleNet~\cite{szegedy2016inception,szegedy2015going,szegedy2015rethinking} introduced an Inception layer for the composition of various receptive fields. The residual network~\cite{he2016deep,he2016identity,zagoruyko2016wide}, which adds shortcut connections to implement identity mapping, allows more layers to be stacked without running into the gradient vanishing problem. Recent research on CNNs has mostly focused on composing layers rather than the convolution itself.

Other basic units, such as activation and pooling units, have been studied with many variations. Sigmoid~\cite{han1995influence} and tanh were the basic activations for the very first neural network. The rectified linear unit (ReLU)~\cite{nair2010rectified} was suggested to overcome the gradient vanishing problem, and achieved good results without pre-training. Since then, many variants of ReLUs has been suggested, such as the leaky ReLU (LReLU)~\cite{maas2013rectifier}, randomized LReLU~\cite{xu2015empirical}, parametric ReLU ~\cite{he2015delving}, and exponential linear unit~\cite{ClevertUH15}. Other types of activation units have been suggested to learn subnetworks, such as Maxout~\cite{goodfellow2013maxout} and local winner-take-all~\cite{srivastava2013compete}.

Pooling is another basic operation in a CNN to reduce the resolution and enable translation invariance. Max and average pooling are the most popular methods. Spatial pyramid pooling~\cite{he2014spatial} was introduced to deal with inputs of varying resolution. The ROI pooling method was used to speed up detection~\cite{girshick2015fast}. Recently, fractional pooling~\cite{graham2014fractional} has been applied to image classification. Lee et al.~\cite{lee2016generalizing} proposed a general pooling method that combines pooling and convolution units. On the other hand, Springenberg et al.~\cite{Springenberg2015} showed that using only convolution units is sufficient without any pooling.

However, only a few studies have considered convolution units themselves. Dilated convolution~\cite{chen2014semantic,YuKoltun2016} has been suggested for dense prediction of segmentation. It reduces post-processing to enhance the resolution of the segmented result. Permutohedral lattice convolution~\cite{kiefel2014permutohedral} is used to expand the convolved dimension from the spatial domain to the color domain. It enables pairwise potentials to be learned for conditional random fields.

In this paper, we propose a new convolution unit. Unlike conventional convolution and its variants, this unit does not have a fixed shape of the receptive field, and can be used to take more diverse forms of receptive fields for convolutions. Moreover, its shape can be learned during the training procedure. Since the shape of the unit is deformable and learnable, we call it the active convolution unit (ACU).

The main contribution of this paper is this convolution unit to provide greater flexibility and representation power to a CNN for a meaningful improvement in the image classification benchmark. We explain the new convolution unit in Section~\ref{sec:ACU}. In Sections~\ref{sec:exp.plain} and~\ref{sec:exp.res}, we report experiments of our unit on plain and residual networks. Finally, we show the results on a general dataset in Section~\ref{sec:exp.general}, and conclude the paper in Section~\ref{sec:conclusion}.

\section{Active Convolution Unit} \label{sec:ACU}
In this section, we describe the basic concept of the ACU. The ACU is a new type of convolution unit with position parameters. Conventionally, the shape of a convolution unit is fixed when the network is initialized, and is not flexible. Fig.~\ref{fig:ACU shape} shows the concept of the ACU. The ACU can define more diverse forms of the receptive fields for convolutions with learnable positions parameters. Inspired by the nervous system, we call one acceptor of the ACU the \textit{synapse}. Position parameters can be differentiated, and the shape can be learned through backpropagation.

\begin{figure}
\centering
\includegraphics[width=0.26\textwidth]{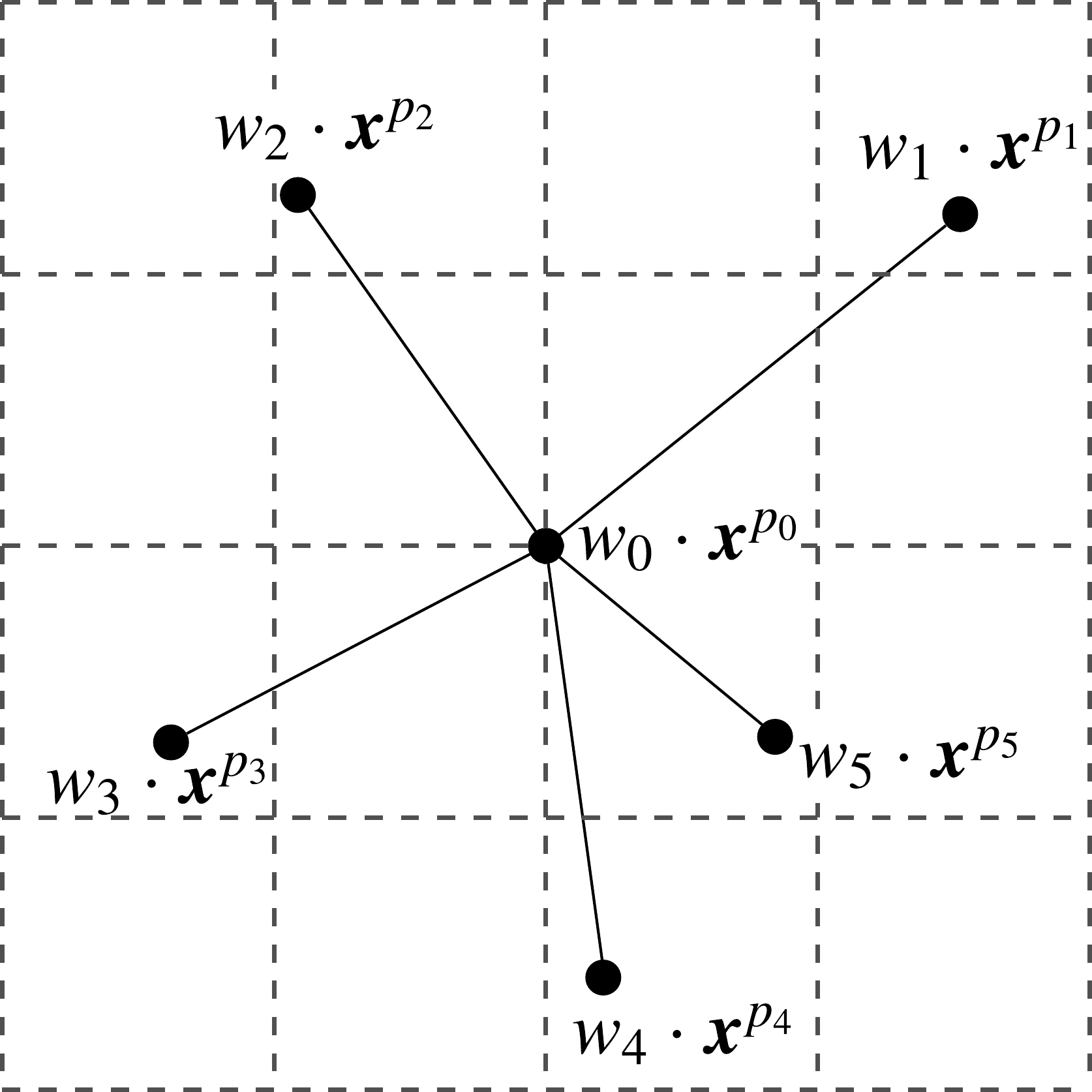}
\caption{Concept of the ACU. Black dots represent each synapse. The ACU’s output is the summation of values in all positions $p_k$ multiplied by weight. The position is parameterized by $p_k$.}
\label{fig:ACU shape}
\end{figure}

\subsection{Capacity of ACU}
The ACU can be considered a generalization of the convolution unit. Any conventional convolution can be represented with the ACU by setting the positions of the synapses properly and fixing all positions. Dilated convolution can be also represented by multiplying the dilation factor with the position parameters. Compared to a conventional convolution, the ACU can generate fractional dilated convolutions and be used to directly calculate the results of the interpolated convolution. It can also be used to define $K$ synapses without any restriction (e.g., cross-shaped convolution with five synapses, or a circular convolution with many synapses).

At the network level, the ACU converts a discrete input space to a continuous one (Fig.~\ref{fig:ACU_concept}). Since the ACU uses bilinear interpolation between adjacent neurons, synapses can connect inter-neuron spaces. This lends greater representational power to the convolution units. The position parameters control the synapses that connect neuron spaces, and the synapses can move around the neuron space to reduce error.

\begin{figure}
	\centering	
	\begin{tabular}{ccc}
		\includegraphics[width=0.20\textwidth]{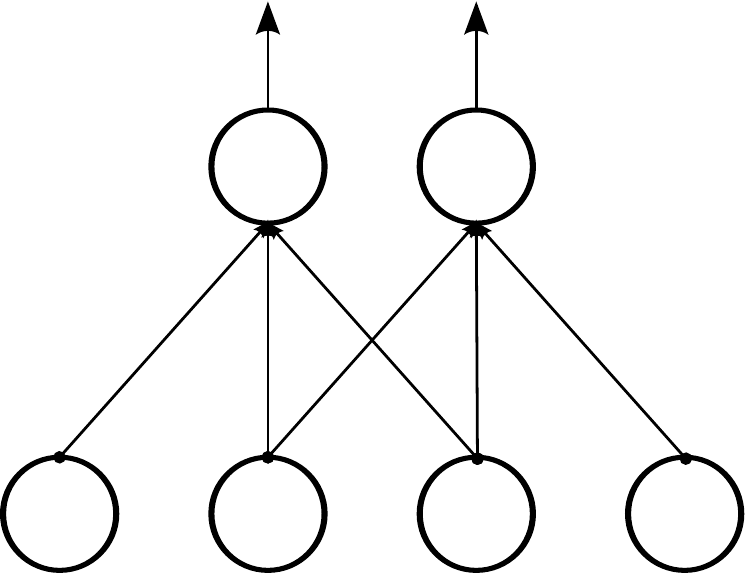} &
		\includegraphics[width=0.22\textwidth]{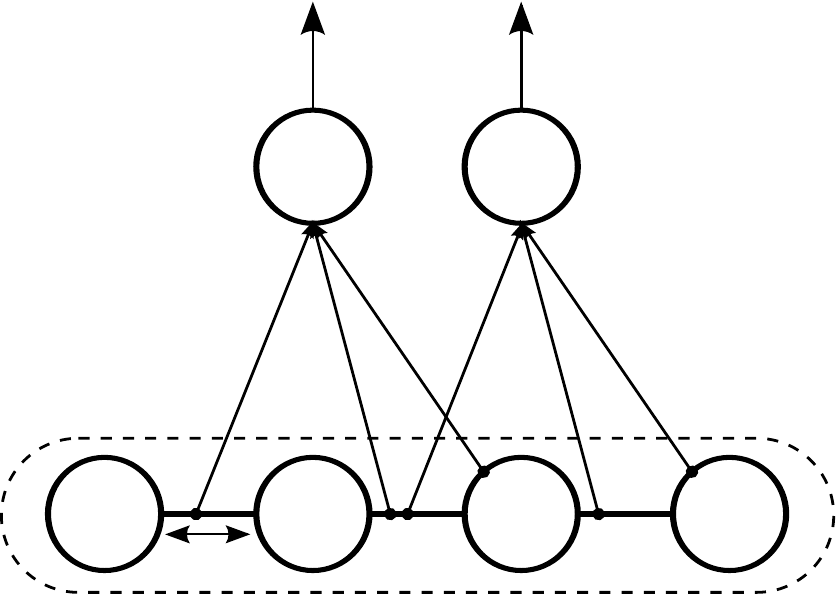} \\
		(a) Basic convolution & (b) Active convolution
	\end{tabular}	
	\caption{Comparison of a conventional convolution unit with the ACU. (a) Conventional convolution unit with four input neurons and two output neurons. (b) Unlike the convolution unit, the synapses of the ACU can be connected at inter-neuron positions and are movable.}\label{fig:ACU_concept}
\end{figure}

\subsection{Formulation}
A convolution unit has a number of learnable filters, and each filter is convolved with its receptive field. A filter of the convolution unit has weight $\boldsymbol{W}$ and bias $\boldsymbol{b}$. For simplicity, we provide a formulation for the case with one output channel, but we can simply expand this formulation for cases with multiple output channels. A conventional convolution can be formulated as shown in Eqs.~\eqref{eq:conv eq1} and \eqref{eq:conv eq2}:

\begin{equation}
\label{eq:conv eq1}
\boldsymbol{Y} = \boldsymbol{W} * \boldsymbol{X} + \boldsymbol{b}
\end{equation}
\begin{equation}
\label{eq:conv eq2}
y_{m,n} = \sum_{c}\sum_{i,j} w_{c,i,j} \cdot x_{c,m+i,n+j} + b
\end{equation}
where $c$ is the index of the input channel and $b$ the bias. $m$ and $n$ are the spatial positions, and $w_{c,i,j}$ and ${x}_{c,m,n}$ are the weight of the convolution filter and the value in the given channel and position, respectively.

Compared with the basic convolution, the ACU has a learnable position parameter $\theta_p$, which is the set of positions of the synapses (Eq.~\eqref{eq:position param}):

\begin{equation}
\label{eq:position param}
\theta_p = \{p_k|0\leq k < K\}
\end{equation}
where $k$ is the index of the synapse and $p_k = (\alpha_k, \beta_k) \in \mathbb{R}^2$. The parameters $\alpha_k$ and $\beta_k$ define the horizontal and vertical displacements, respectively, of the synapse from the origin. With parameter $\theta_p$, the ACU can be defined as given in Eqs.~\eqref{eq:ACU eq1},\eqref{eq:ACU eq2}:

\begin{equation}
\label{eq:ACU eq1}
\boldsymbol{Y} = \boldsymbol{W} * \boldsymbol{X}_{\theta_p} + \boldsymbol{b}
\end{equation}
\begin{equation}
\label{eq:ACU eq2}
\begin{aligned}
y_{m,n} = &\sum_{c}\sum_{k} w_{c,k} \cdot {x}^{p_k}_{c,m,n} + b\\
= &\sum_{c}\sum_{k} w_{c,k} \cdot x_{c,m+\alpha_k,n+\beta_k} + b
\end{aligned}
\end{equation}

For instance, the conventional $3 \times 3$ convolution can be represented by the ACU with $\theta_p = \{(-1,-1), (0,-1), $ $(1,-1),(-1,0),(0,0),(1,0),(-1,1),(0,1),(1,1)\}$.

In this paper, $\theta_p$ is shared across all output units $y_{m,n}$. If the number of synapses, input channels, and output channels are $K$, $C$, and $D$, respectively, the dimensions of weight $\boldsymbol{W}$ should be $D \times C \times K$. The additional parameters for the ACU are $2\times K$; this number is very small compared to the number of weight parameters.

\subsection{Forward Pass}
Because the position parameter $p_k$ is a real number, $x_{c,m+\alpha_k,n+\beta_k}$ can refer to a nonlattice point. To obtain the value of a fractional position, we use bilinear interpolation:

\begin{figure}
	\centering
	\includegraphics[width=0.40\textwidth]{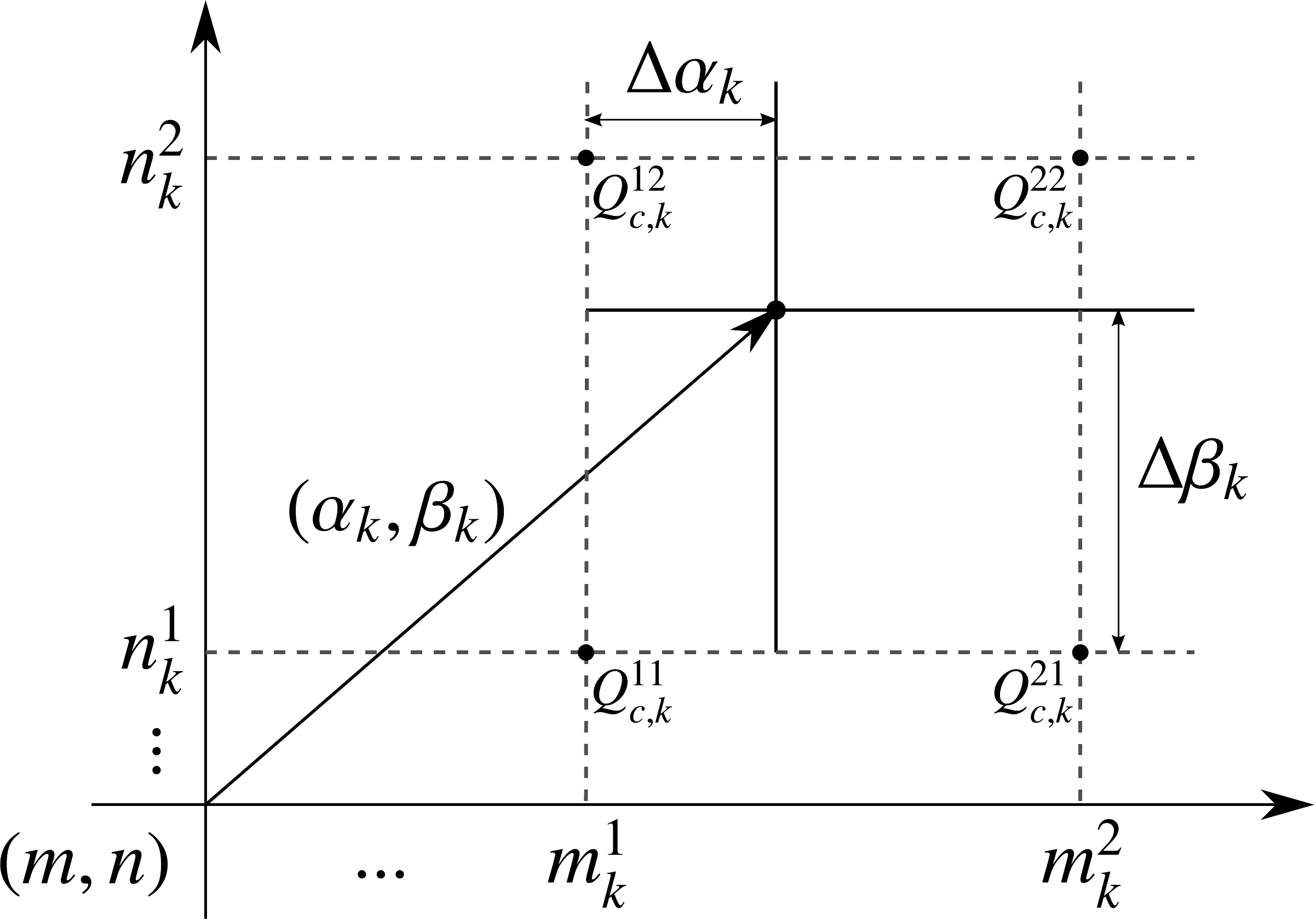}
	\caption{Coordinate system of interpolation. $m,n$ represent the base position of the convolution $\alpha_k, and \beta_k$ is the displacement of the $k$th synapse.}
	\label{fig:interpolation}
\end{figure}

\begin{equation}
\label{eq:interpolation}
\begin{aligned}
{x}^{p_k}_{c,m,n} = &x_{c,m + \alpha_k, n + \beta_k}\\
= &\{Q^{11}_{c,k}\cdot(1 - \Delta\alpha_k)\cdot(1 - \Delta\beta_k)\\
&+Q^{21}_{c,k}\cdot\Delta\alpha_k\cdot(1- \Delta\beta_k)\\
&+Q^{12}_{c,k}\cdot(1-\Delta\alpha_k)\cdot\Delta\beta_k\\
&+Q^{22}_{c,k}\cdot\Delta\alpha_k\cdot\Delta\beta_k\}
\end{aligned}
\end{equation}
where
\begin{equation}
\label{eq:delta alpha,beta}
\begin{aligned}
\Delta\alpha_k = \alpha_k - \lfloor	{\alpha_k}\rfloor\\
\Delta\beta_k = \beta_k - \lfloor	{\beta_k}\rfloor
\end{aligned}
\end{equation}

\begin{equation}
\label{eq:m1,n1 def}
\begin{aligned}
m^1_k = m + \lfloor	{\alpha_k}\rfloor, m^2_k = m^1_k + 1,\\
n^1_k = n + \lfloor	{\beta_k}\rfloor, n^2_k = n^1_k + 1
\end{aligned}
\end{equation}

\begin{equation}
\label{eq:Q def}
\begin{aligned}
Q^{11}_{c,k} = x_{c,m^1_k,n^1_k}, Q^{12}_{c,k} = x_{c,m^1_k,n^2_k},\\
Q^{21}_{c,k} = x_{c,m^2_k,n^1_k}, Q^{22}_{c,k} = x_{c,m^2_k,n^2_k}.
\end{aligned}
\end{equation}

We can obtain the value of the fractional position by using the four nearest integer points $Q^{ab}_{c,k}$. The interpolated value ${x}^{p_k}_{c,m,n}$ is continuous with respect to any $p_k$, even at the lattice point.

\subsection{Backward Pass}
The ACU has three types of parameters: weight, bias, and position. They can all be differentiated and learned thorough backpropagation. The partial derivative of weight $w$ is the same as that for a conventional convolution, except that the interpolated value of ${x}^{p_k}_{c,m,n}$ (Eq.~\eqref{eq:weight diff}) is used. The gradient of the bias is the same as that of the conventional convolution (Eq.~\eqref{eq:bias diff}):

\begin{equation}
\label{eq:weight diff}
\frac{\partial y_{m,n} }{\partial w_{c,k}} = {x}^{p_k}_{c,m,n}~,
\end{equation}
\begin{equation}
\label{eq:bias diff}
\frac{\partial y_{m,n} }{\partial b} = 1.
\end{equation}

Because we use bilinear interpolation, the positions are simply differentiated as defined in Eqs.~\eqref{eq:pos diff x},\eqref{eq:pos diff y}:

\begin{equation}
\label{eq:pos diff x}
\begin{aligned}
\frac{\partial y_{m,n} }{\partial \alpha_k}
= \sum_{c} w_{c,k} \cdot &\{(1-\Delta\beta_k)\cdot(Q^{21}_{c,k}-Q^{11}_{c,k})
\\&+ \Delta\beta_k \cdot(Q^{22}_{c,k} - Q^{12}_{c,k})\},
\end{aligned}
\end{equation}
\begin{equation}
\label{eq:pos diff y}
\begin{aligned}
\frac{\partial y_{m,n} }{\partial \beta_k}
= \sum_{c} w_{c,k} \cdot &\{(1-\Delta\alpha_k)\cdot(Q^{12}_{c,k}-Q^{11}_{c,k})
\\&+ (\Delta\alpha_k \cdot(Q^{22}_{c,k} - Q^{21}_{c,k})\}.
\end{aligned}
\end{equation}

The derivatives of the positions are valid in subpixel regions, i.e., between lattice points. They may not be differentiable at lattice points because we use the four nearest neighbors for interpolation. Empirically, however, we found that this was not a problem if we set an appropriate learning rate for the position parameters. This is explained in Section~\ref{subsec:norm grad}.

The differential with respect to the input of the ACU is given simply as 

\begin{equation}
\label{eq:bottom diff}
\frac{\partial y_{m,n} }{\partial {x}^{p_k}_{c,m,n}}
= w_{c,k}.
\end{equation}

\subsection{Normalized Gradient} \label{subsec:norm grad}
The backpropagated value of the position of the synapse controls its movement. If the value is too small, the synapse stays in almost the same position and the ACU has no effect. By contrast, a large value diversifies the synapses. Hence, controlling the magnitude of movement is important. The partial derivatives with respect to position are dependent on the weight, and the backpropagated error can fluctuate across layers. Hence, determining the learning rate of the position is difficult.

One way to reduce the fluctuation of the gradient across layers is to use only the direction of the derivatives, and not the magnitude. When we use the normalized gradient of position, we can easily control the magnitude of movement. We observed in experiments that the use of a normalized gradient made it easier to train and achieve a good result. The normalized gradient of the position is defined as

\begin{equation}
\label{eq:norm grad}
\begin{aligned}
\overline{\frac{\partial L }{\partial \alpha_k}} = \frac{\partial L }{\partial \alpha_k} / Z,\\
\overline{\frac{\partial L }{\partial \beta_k}} = \frac{\partial L }{\partial \beta_k} / Z
\end{aligned}
\end{equation}
where $L$ is loss function, and $Z$ is the normalization factor:
\begin{equation}
\label{eq:norm factor}
Z = \sqrt{(\frac{\partial L }{\partial \alpha_k})^ 2 + (\frac{\partial L }{\partial \beta_k})^2}
\end{equation}

An initial learning rate of 0.001 is typically appropriate, which means that a synapse moves 0.001 pixel in an iteration. In other words, a synapse can move a maximum of one pixel in 1,000 iterations.

\subsection{Warming up}
As noted above, the direction of movement is influenced by weight. Since the weight is initialized with a random distribution, the movement of the synapse wanders randomly in early iterations. This can cause the position to stick to a local minimum; hence, we first warm up the network without learning the position parameters. In early iterations, the network only learns weights with fixed shape. It then learns positions and weights concurrently. This helps the synapses learn a more stable shape; in experiments, we obtained an improvement of 0.1\%--0.2\% over benchmark data.

\subsection{Miscellaneous}
The index of synapse $k$ starts at $0$, which represents the base position. In experiments, we fixed $p_0$ to the origin, where $p_0 = (\alpha_0, \beta_0) = (0,0)$, ${x}^{p_0}_{c,m,n} = x_{c,m,n}$, and the learning rate for $p_0$ was set to $0$. Since $p_0$ was fixed, we did not need to assign a parameter to it, but we set the index to 0 for convenience.

The ACU was implemented on Caffe~\cite{jia2014caffe}. Owing to the calculation of bilinear interpolations, the ACU was approximately 1.6 times slower than the conventional convolution for the forward pass with a normal graphics processing unit (GPU) implementation. If we implement the ACU using a dedicated multi-processing API, like the CUDA Deep Neural Network library (cuDNN)~\cite{chetlur2014cudnn}, it can be run faster than the current implementation.

\section{ACU with a Plain Network} \label{sec:exp.plain}

To evaluate the performance of the proposed ACU, we created a simple network consisting of convolution units without a pooling layer~\cite{Springenberg2015} (Table~\ref{table:baseline}). Batch normalization~\cite{ioffe2015batch} and ReLU were applied after all convolutions.

\begin{table}
	\begin{center}
		\begin{tabular}{c|c|c}
		\hline
		\textbf{Layer Name}	& \textbf{Type}	& \textbf{Size} \\	\hline \hline
		conv0		& 1$\times$1 conv	& 16\\	\hline
		conv1/1	& 3$\times$3 conv	& 48\\
		conv1/2	& 3$\times$3 conv	& 48\\	\hline
		conv2/1	& 3$\times$3 conv,stride 2	& 96\\
		conv2/2	& 3$\times$3 conv	& 96\\	\hline
		conv3/1	& 3$\times$3 conv,stride 2	& 192\\
		conv3/2	& 3$\times$3 conv	& 192\\	\hline
		fc1	& 1$\times$1 conv	& 1000\\	\hline
		-	& global average-pooling	&\\	\hline
		-	& softmax 10	&\\	\hline
		\end{tabular}
	\end{center}
	\caption{Baseline structure of the plain network.}
	\label{table:baseline}
\end{table}

We used the CIFAR dataset for the experiment~\cite{krizhevsky2009learning}. CIFAR-10/100 is widely used for image classification, and consists of 60k $32 \times 32$ color images: 50k for training and 10k for testing in 10 and 100 classes, respectively. The weight parameters were initialized by using the method proposed by He et al.~\cite{he2015delving} for all experiments. The positions of the synapses were initialized with the shape of a conventional $3 \times 3$ convolution.

\subsection{Experimental results}
We simply changed all $3 \times 3$ convolution units to ACUs from the baseline. The $3 \times 3$ convolution unit can be substituted by an ACU with nine synapses (one is fixed: $p_0$). The position parameters were shared across all kernels in the same layer; thus, $8 \times 2$ more parameters were required per ACU layer. We replaced six convolution layers with the ACU; hence, only 96 parameters were added compared to the baseline. To test the performance only of the ACU, all other parameters were kept constant.

The network was trained with 64k iterations by using stochastic gradient descent with the Nesterov momentum. The initial learning rate was 0.1, and was divided by 10 after 32k and 48k iterations. The batch size was 64, and the momentum was 0.9. We used L2 regularization, and the weight decay was 0.0001. We warmed up the network for 10k iterations without learning positions. We used the normalized gradient for the synapses, and the learning rate was set to 0.01 times the base learning rate. When the base learning rate decreased, so did the learning rate of the synapse, which meant that their movement was limited over iterations.

We preprocessed input data with commonly used methods~\cite{goodfellow2013maxout}: global contrast normalization and ZCA whitening. For data augmentation, we performed random cropping on images padded by four pixels on each side as well as horizontal flipping, as in previous studies~\cite{goodfellow2013maxout}.

\begin{table}
	\begin{center}
		\begin{tabular}{c|c|c}
			\hline 			
			\textbf{Network}  & \textbf{CIFAR-10(\%)} & \textbf{CIFAR-100(\%)}  \\ 		\hline 				\hline 
			baseline & 8.01    & 27.85\\			
			ACU      & 7.33    & 27.11\\			\hline
			Improvement & \textbf{+0.68}& \textbf{+0.74}\\
		\end{tabular} 
	\end{center}	
	\caption{Error rate on test set with the plain network. Using the ACU on the plain network improved accuracy over the baseline.}
	\label{table:base exp result}	
\end{table}	

Table~\ref{table:base exp result} presents the experimental results. The baseline network obtained error rates of 8.01\% with CIFAR-10 and 27.85\% with CIFAR-100. Once we changed the convolution units to ACUs, the error rate dropped to 7.33\% with CIFAR-10, a 0.68\% improvement over the baseline. We obtained a similar result with CIFAR-100: the ACU reduced the error rate by 0.74\%. Fig.~\ref{fig:acu_loss_graph} shows the training curve for CIFAR-10. Both the training loss and the test error were less than those of the baseline.

\begin{figure}
	\centering
	\includegraphics[width=0.48\textwidth]{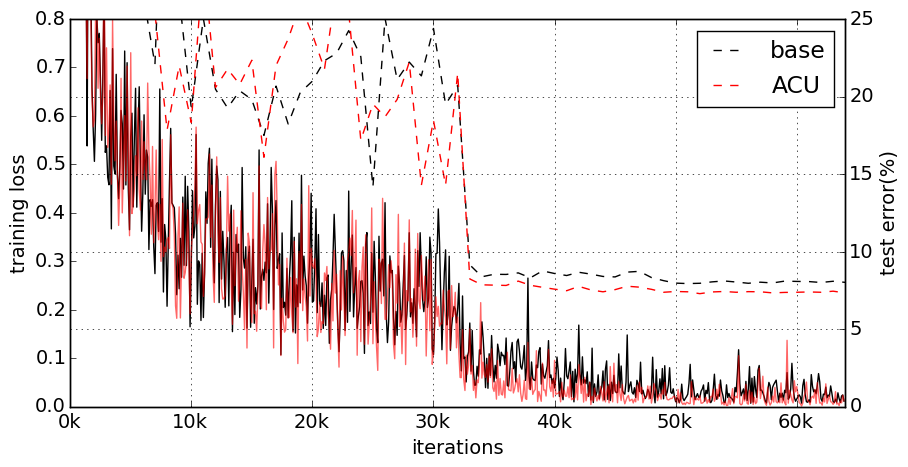}
	\caption{Training curve for CIFAR-10 with the plain network. Solid lines represent training loss and the dotted line the test error.}
	\label{fig:acu_loss_graph}
\end{figure}

\subsection{Learned Position}

Fig.~\ref{fig:ACU learned filter} shows the results of position learning. The exact positions were different in each trial due to the weight initialization but had similar characteristics. In the lower layer, the ACU settled down to a conventional $3 \times 3$ convolution. By contrast, the upper layers tended to grow. The last ACU was similar to the two dilated convolutions. Since we normalized the magnitude of movement, all synapses moved by the same magnitude in an iteration. Thus, this phenomenon was not due to gradient flow. We think that the lower layers focused on capturing local features and the higher layers became wider to integrate the spatial features from the preceding layers.

Fig.~\ref{fig:displacement} shows the movement of $\alpha_k$ in the first and the last layers. We see the first layer tended to maintain the initial shape of the receptive field. Unlike in the first layer, the positions of synapses monotonically grew in the last layer.

Although the lower layer was similar to a conventional convolution, we found that the ACU was still effective for it. Since the synapses move continuously during training, the same inputs and weights do not yield the same outputs. The output of the ACU is spatially deformed in every iteration, which lends an augmentation effect without explicit data augmentation.

\begin{figure}
	\centering
	\includegraphics[width=0.48\textwidth]{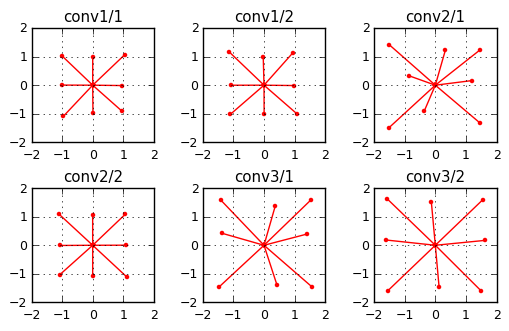}
	\caption{Learned position on CIFAR-10. High layers tended to grow, and the first layer was almost identical to the conventional convolution.}
	\label{fig:ACU learned filter}
\end{figure}
\begin{figure}
	\centering
	\includegraphics[width=0.45\textwidth]{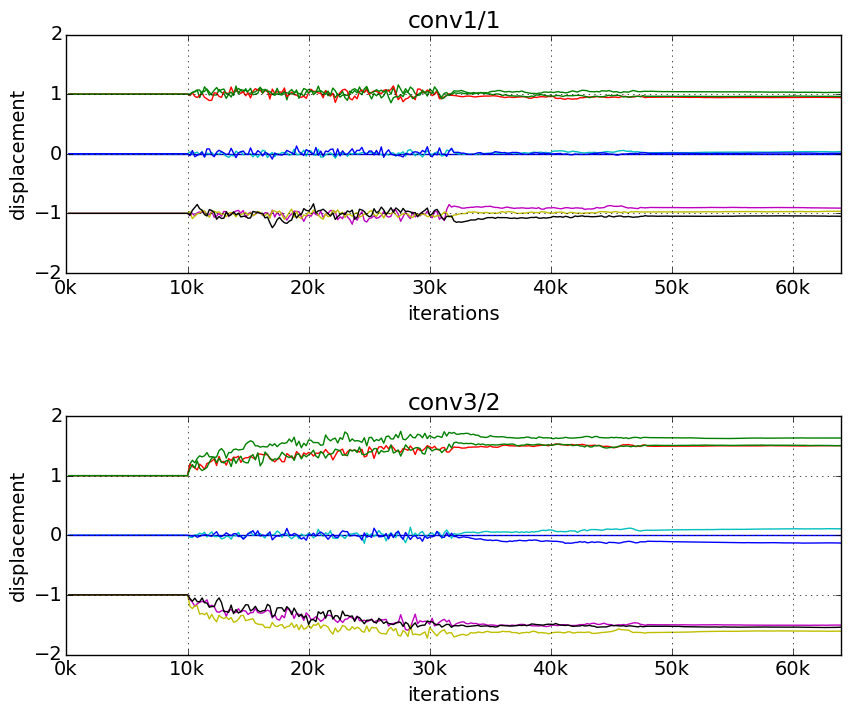}
	\caption{Movement of position. Each figure shows changes in the horizontal position ($\alpha_k$) with iterations. Each figure shows the first and the last ACU layers. Each color represents one synapse (9 synapses in total).}
	\label{fig:displacement}
\end{figure}

\section{ACU with the Residual Network} \label{sec:exp.res}

Recent work has shown that a residual structure~\cite{he2016deep,he2016identity,szegedy2016inception,zagoruyko2016wide} can improve performance using very deep layers. We experimented on our convolution unit with a residual network and investigated how the ACU can collaborate with residual architecture. All our experiments were based on pre-activation~\cite{he2016identity} residual, and the experimental parameters were the same as those described in the previous section, except for network structure.

\begin{table}
	\begin{center}
		\begin{tabular}{c|c|c}
			\hline 	
			\textbf{Network} & \textbf{CIFAR-10(\%)} & \textbf{CIFAR-100(\%)} \\ 
			\hline 	
			\hline 							
			Basic residual 	& 8.01 				& 30.06	 \\		\hline 				
			ACU 			&  7.54 			& 29.38	 \\		\hline 				
			Improvement 	& \textbf{+0.47}	& \textbf{+0.68}\\		
			\hline	\hline
			Bottleneck residual	& 7.64  	& 27.93 	\\		\hline 				
			ACU 				&  7.12	 	&  27.47	\\		\hline 							
			Improvement 	& \textbf{+0.52}	& \textbf{+0.46}\\				
		\end{tabular} 
	\end{center}
	\caption{Error rate on test set with a residual network. Using the ACU with the residual network improved accuracy over the baseline.}
	\label{table:res exp result}	
\end{table}	

\begin{figure}
	\centering
	\begin{tabular}{ccc}
	\includegraphics[width=0.13\textwidth]{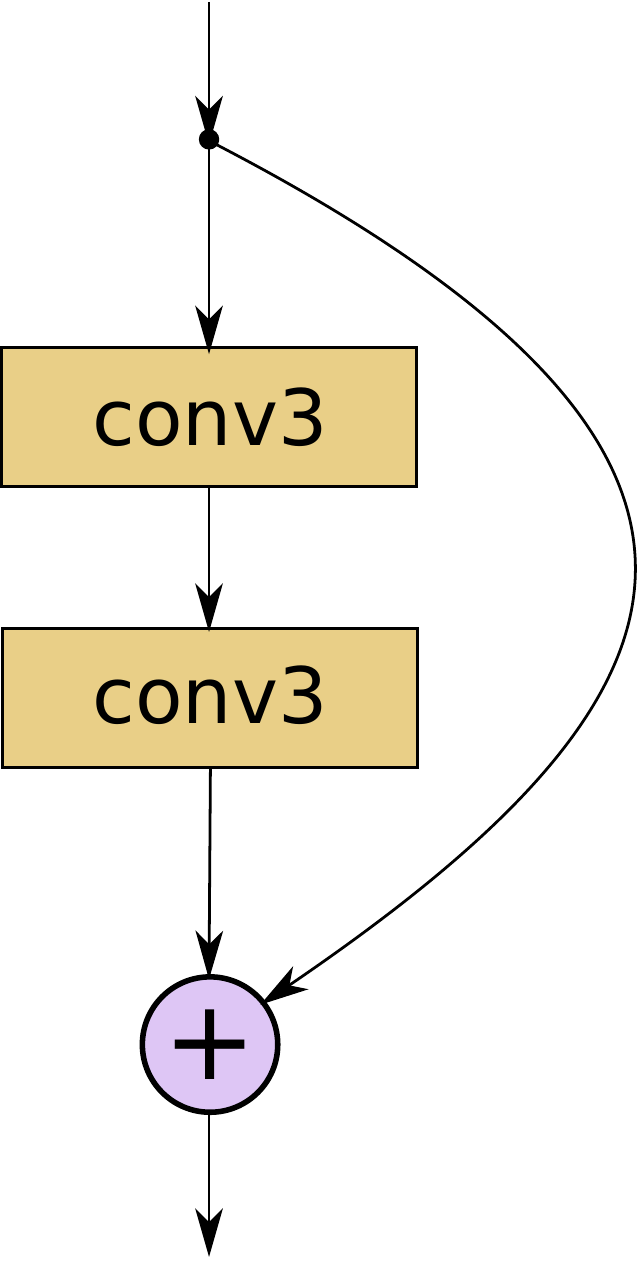} &
	\includegraphics[width=0.17\textwidth]{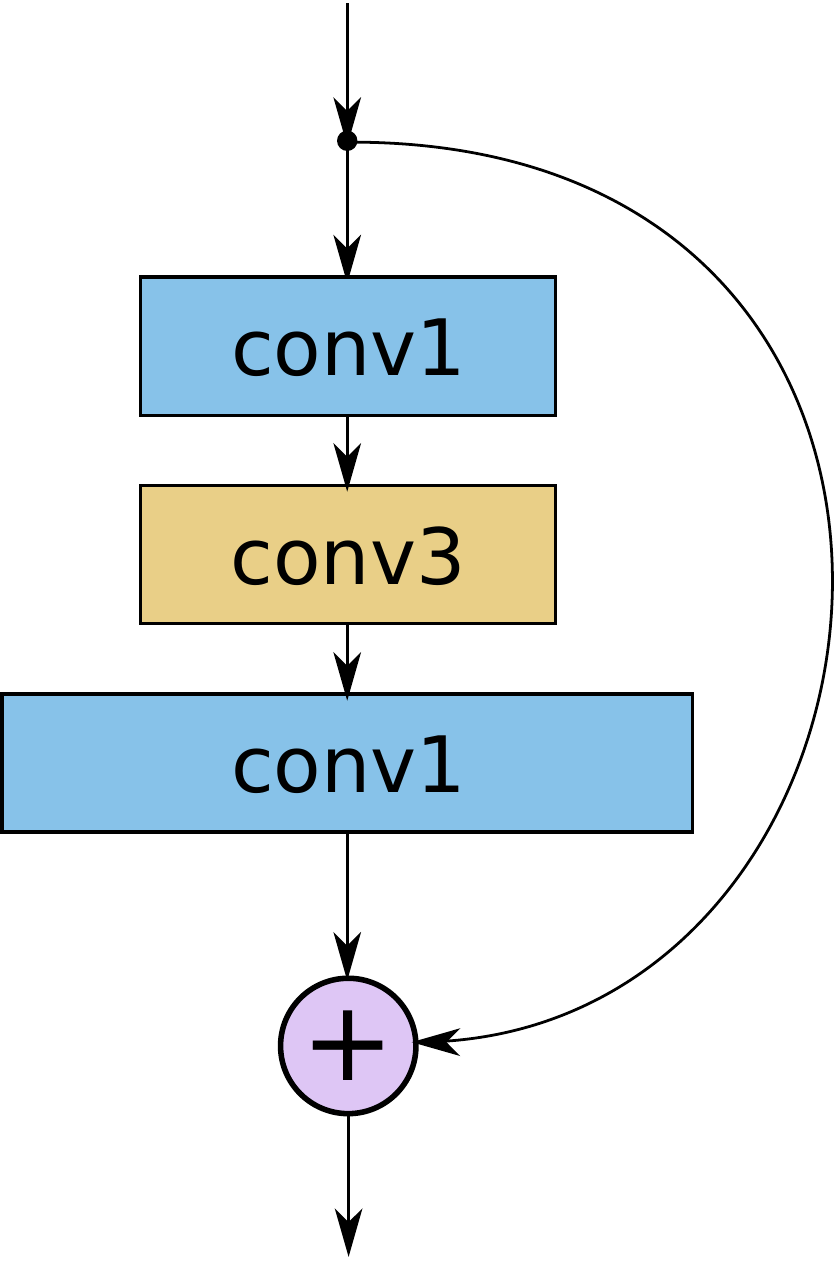} \\
	(a) Basic & (b) Bottleneck
	\end{tabular}
	\caption{Comparison of residual structures: (a) basic residual and (b) bottleneck structure. We changed all \textit{conv3} layers to ACUs.}
	\label{fig:bottleneck struct}
\end{figure}

\subsection{Basic Residual Network}
The basic residual network consists of residual blocks with two $3 \times 3$ convolutions in series. To check the availability of the ACU, we formed a 32-layer residual network with five residual blocks. We used projection shortcuts to increase the number of dimensions, where the other shortcuts were identity.

Table~\ref{table:res exp result} presents the experimental results. The basic residual network achieved an 8.01\% error rate with CIFAR-10. We replaced all convolution units with ACUs in the residual blocks. This yielded a 7.54\% error, an improvement of 0.47\% over the baseline. With CIFAR-100, we obtained a 0.68\% improvement, which was better than that with CIFAR-10.

\subsection{Bottleneck Residual Network}
A bottleneck residual unit consists of three successive convolution layers: $1 \times 1$, $3 \times 3$, and $1 \times 1$~\cite{he2016deep, he2016identity}. The first layer reduces the number of dimensions and the last $1 \times 1$ convolution restores the original dimensions. Fig.~\ref{fig:bottleneck struct} shows the difference between the basic and the bottleneck residual blocks. This architecture has a similar complexity to that of the basic residual block, but can increase network depth.

The 32-layer basic residual network was converted into a 47-layer bottleneck network. The number of parameters was nearly the same, and this network achieved a 7.64\% error rate (Table~\ref{table:res exp result}). Since the bottleneck block had only one $3 \times 3$ convolution, we only changed one convolution to ACU. We obtained a 7.12\% error with CIFAR-10, a 0.52\% improvement over the bottleneck baseline. We obtained a similar result (0.46\% gain) with CIFAR-100. The training loss and the test error are shown in Fig.~\ref{fig:acu_loss_graph_res}

\begin{figure}
	\centering
	\includegraphics[width=0.48\textwidth]{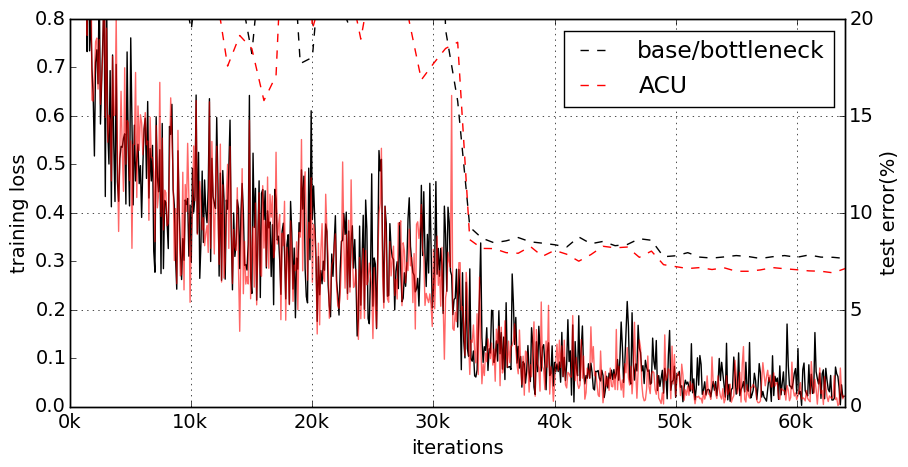}
	\caption{Training curve for CIFAR-10 with the residual network. Solid lines represent training loss and the dotted line the test error.}
	\label{fig:acu_loss_graph_res}
\end{figure}

\begin{figure}
	\centering
	\includegraphics[width=0.47\textwidth]{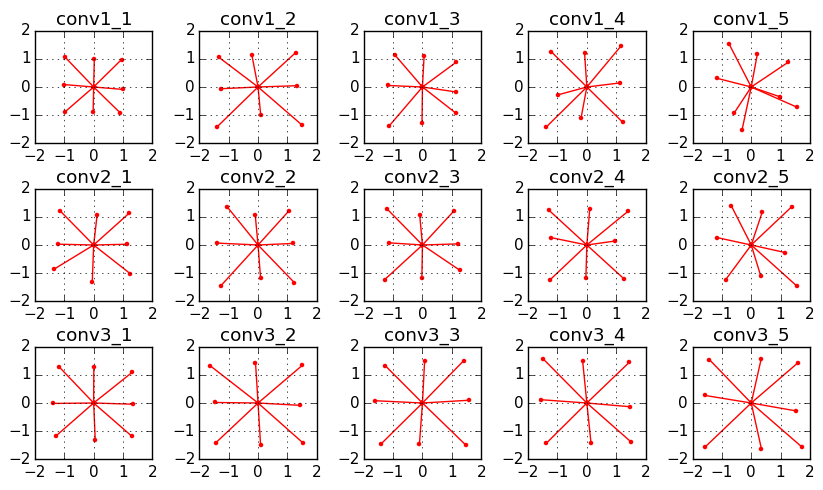}
	\caption{Learned positions with the bottleneck residual network. Each row represents ACUs in each residual block with the same number of dimensions.}
	\label{fig:res learned filter}
\end{figure}

\subsection{Learned Position}

Fig.~\ref{fig:res learned filter} shows the results of the learned filter for the bottleneck structure. In total, 15 convolutions were changed to ACUs. Each row shows the learned position of the ACU in the residual block with the same number of dimensions. As in the plain network, the positions of the synapses became wider at higher layers. ACUs in different residual blocks with the same number of dimensions tended to learn different shapes of the receptive field. This shows that each ACU extracted unique features with different views.

\section{Experiment on Place365} \label{sec:exp.general}

In the previous experiment, we had used the CIFAR-10/100 datasets to check the validity of our approach. However these sets had consisted of small image patches of limited data size. To verify our approach in more realistic situations, we tested the ACU with the Place365-Standard benchmark set~\cite{zhou2016places,zhou2014learning}.

Place365 is the latest subset of the Place2 database, which contains more than 10 million images. Place365-Standard is a benchmark set with more than 1.8 million training images in 365 scene categories. 

To represent plain and residual networks, we trained AlexNet~\cite{Krizhevsky2012} and a 26-layer residual network with a bottleneck structure. We resized images to $256 \times 256$ pixels, and randomly cropped and flipped them horizontally for augmentation. The networks were trained for 400k iterations by using stochastic gradient descent. We divided the learning rate by 10 after 200k and 300k iterations, and used four GPUs with 32 batch sizes each.

When we trained the network with the ACU, we warmed it up network with 50k iterations with fixed positions. We tested the networks with a validation set containing 100 images per class. Top-1 and top-5 accuracies are obtained by the standard 10-crop testing~\cite{Krizhevsky2012}.

\subsection{AlexNet}
Following the conventions of AlexNet, we used a weight decay of 0.0005, and the momentum was 0.9. The initial learning rate was 0.01. We could have changed the $11 \times 11$ and $5 \times 5$ convolutions to ACUs, but decided to only change the $3 \times 3$ convolutions because we had not completed analyzed the effects of large convolutions. Since the second and third layers of the $3\times 3$ convolution split the channels by two in AlexNet (Fig.~\ref{fig:alexnet_struct}), we assigned two set of positions to each convolution, and five sets of position parameters were used in total.

Simply by changing the $3 \times 3$ convolution layers to ACUs, we obtained an improvement of 0.79\% for AlexNet (Table~\ref{table:result_place365}). Fig.~\ref{fig:loss_place365} shows the training curve of AlexNet and its ACU network. Their accuracies were almost identical during the warming up iterations; but after warming up (50k iterations), test error began to decrease. This shows the effect of the ACU.

Fig.~\ref{fig:365_position}(a) shows the position learned after training. The first ACU layer (\textit{conv3}) was not the first layer of the network (\textit{conv1}), and the shape of this layer was not similar to that of conventional convolutions, unlike in previous experiments. The learned shape of the synapses is interesting in that it is similar to combinations of two receptive fields of a convolution. With small numbers of weight parameters, the ACU tried to learn multiple receptive field features.

\subsection{Residual Network}
To test with the residual network, we constructed a 26-layer residual network, which was an 18-layer basic network~\cite{he2016deep} converted into a bottleneck structure. Due to restrictions on hardware, we were not able to use deeper architecture. In the residual network, weight decay was 0.0001 and the initial learning rate was 0.1. To prevent overfitting, we stopped training at 350k iterations.

With the residual network, we replaced all $3 \time 3$ convolutions in the residual block, as in the previous experiment. In total, eight convolutions were replaced with ACUs, and thus, 128 parameters were added. With this small number of parameters, we obtained a 0.49\% gain in top-5 accuracy.

On the both of the plain and the residual networks, we obtained a meaningful improvement using the ACU. This results show that the ACU can be applied not only to simple image classification as in CIFAR-10/100, but also to more general problems.

The final shape of the ACU is shown in Fig.~\ref{fig:365_position}(b). As in our previous experiments, the higher layer tended to enlarge its receptive field. However, its coverage was greater than in the previous CIFAR-10/100 experiment. We think that since the image size of Place365 was larger than that of images in the CIFAR dataset, the last layer preferred a larger receptive field.

\begin{figure}
	\centering
	\includegraphics[width=0.48\textwidth]{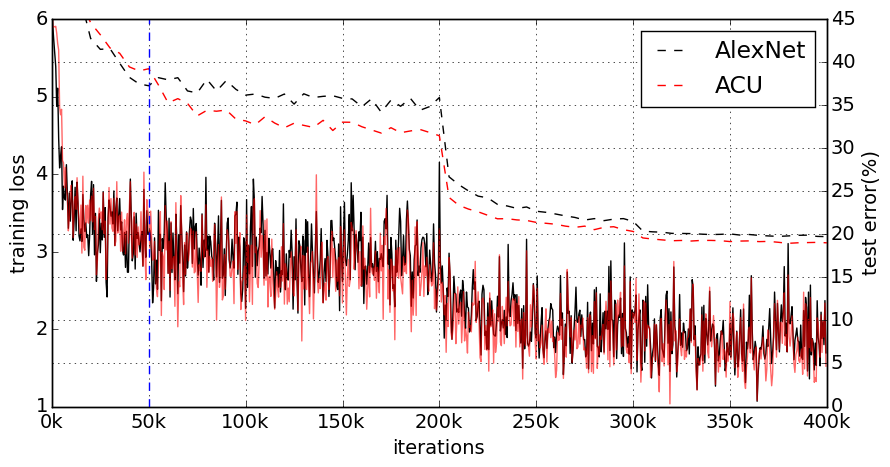}
	\caption{Training curve of AlexNet on Place365. Blue dotted line represents the warming up iteration. After the warm up, test error began to decrease in the ACU network.}
	\label{fig:loss_place365}
\end{figure}

\begin{table*}[t]
	\begin{center}
		\begin{tabular}{c|ccc|ccc}
			\hline 
			& \multicolumn{3}{c|}{\textbf{Top-1(\%)}} & \multicolumn{3}{c}{\textbf{Top-5(\%)}} \\ 
			\textbf{Network} & \textbf{base} & \textbf{ACU} & \textbf{Improvement}& \textbf{base} & \textbf{ACU} & \textbf{Improvement}\\
			\hline 	
			\hline 
			AlexNet  & 51.68 & 52.28 & \textbf{0.6} & 81.29 & 82.08 &\textbf{0.79}\\
			\hline 
			ResNet26 & 55.33 & 55.71 & \textbf{0.38} & 85.24 & 85.73 & \textbf{0.49}\\			
			\hline 									
			
		\end{tabular} 
	\end{center}
	\caption{Classification accuracy on the validation set. We used the average score over standard 10 crops of images.}
	\label{table:result_place365}		
\end{table*}

\section{Conclusion} \label{sec:conclusion}
In this study, we introduced the ACU that contains position parameters to provide more freedom to a conventional convolution and allow its position to be learned through backpropagation. Through various experiments, we showed the effectiveness of our approach. Simply by changing the convolution layers to our unit, we were able to boost the performance of equivalent networks.

Since the shape of the ACU can be freely defined, we believe that it can expand network architectures in a more diverse manner. In this study, we shared only one set of position parameters; it is possible to define other sets of positions in a layer. Using multiple set of positions, the model’s representational power could be expanded. In future work, we plan to investigate variations in ACU in greater detail.

\begin{figure}
	\centering
	\includegraphics[width=0.35\textwidth]{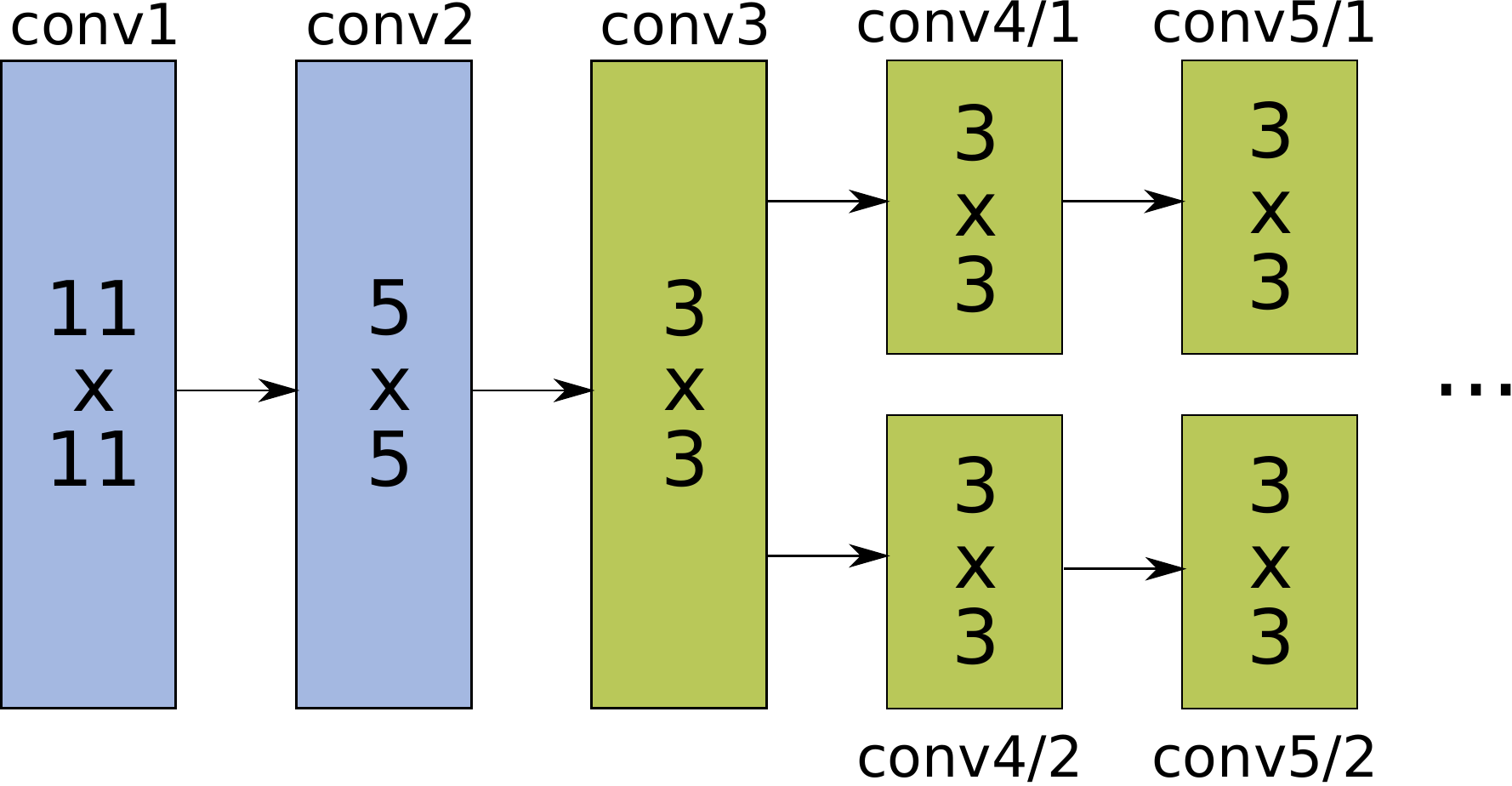}
	\caption{Structure of AlexNet. We changed all $3 \times 3$ convolutions to ACU.}
	\label{fig:alexnet_struct}
\end{figure}

\begin{figure}
	\centering
	\begin{tabular}{c}
		\includegraphics[width=0.45\textwidth]{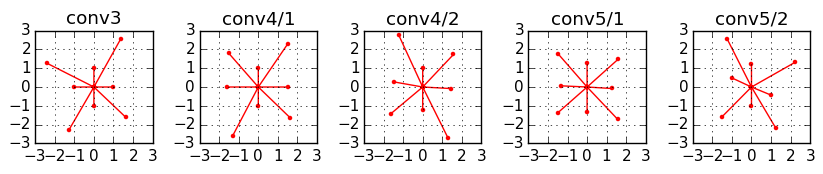} \\(a) AlexNet \\		
		\includegraphics[width=0.45\textwidth]{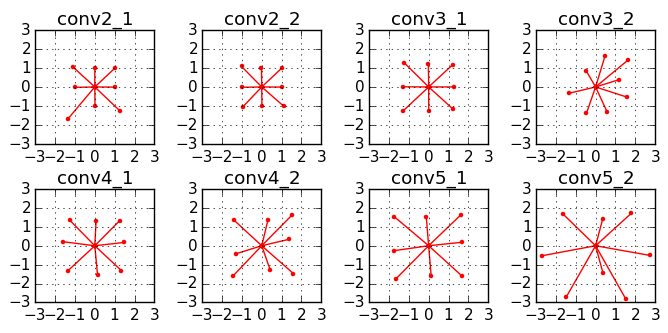} \\(b) Residual network \\
	\end{tabular}
	\caption{Learned positions on Place365 dataset: (a) Result of AlexNet. (b) Result of residual network.}
	\label{fig:365_position}
\end{figure}

{\small
\bibliographystyle{ieee}
\bibliography{egbib}
}

\end{document}